\documentclass[a4paper, 10pt, conference]{ieeeconf}      

\usepackage{FG2026}
\usepackage{graphicx}
\usepackage{makecell}
\usepackage{multirow}
\usepackage{balance}
\usepackage{amsmath}
\usepackage{booktabs}
\usepackage{comment}
\usepackage[colorlinks=true,linkcolor=blue,citecolor=blue]{hyperref}

\IEEEoverridecommandlockouts                              
\def\FGPaperID{****} 

\title{\LARGE \bf
MEGC2026: Micro-Expression Grand Challenge on Visual Question Answering
}

\author{\parbox{16cm}{\centering
    {\large Xinqi Fan$^1$, Jingting Li$^2$, John See$^3$, Moi Hoon Yap$^1$, Su-Jing Wang$^2$, Adrian K. Davison$^1$}\\
    {\normalsize
    $^1$ Department of Computing and Mathematics, Manchester Metropolitan University\\
    $^2$ State Key Laboratory of Cognitive Science and Mental Health, Institute of Psychology, CAS \& \\ Department of Psychology, University of the Chinese Academy of Sciences\\
    $^3$ School of Mathematical and Computer Sciences, Heriot-Watt University Malaysia
    }}
}

\begin{document}

\FGfinaltrue

\ifFGfinal
\thispagestyle{empty}
\pagestyle{empty}
\else
\author{Anonymous FG2026 submission\\ Paper ID \FGPaperID \\}
\pagestyle{plain}
\fi

\maketitle

\begin{abstract}
    Facial micro-expressions (MEs) are involuntary movements of the face that occur spontaneously when a person experiences an emotion but attempts to suppress or repress the facial expression, typically found in a high-stakes environment. In recent years, substantial advancements have been made in the areas of ME recognition, spotting, and generation. The emergence of multimodal large language models (MLLMs) and large vision-language models (LVLMs) offers promising new avenues for enhancing ME analysis through their powerful multimodal reasoning capabilities. The ME grand challenge (MEGC) 2026 introduces two tasks that reflect these evolving research directions: (1) ME video question answering (ME-VQA), which explores ME understanding through visual question answering on relatively short video sequences, leveraging MLLMs or LVLMs to address diverse question types related to MEs; and (2) ME long-video question answering (ME-LVQA), which extends VQA to long-duration video sequences in realistic settings, requiring models to handle temporal reasoning and subtle micro-expression detection across extended time periods. All participating algorithms are required to submit their results on a public leaderboard. More details are available at \url{https://megc2026.github.io}.
\end{abstract}

\section{Introduction}
\begin{figure*}[!t]
    \centering
    \includegraphics[width=0.8\textwidth]{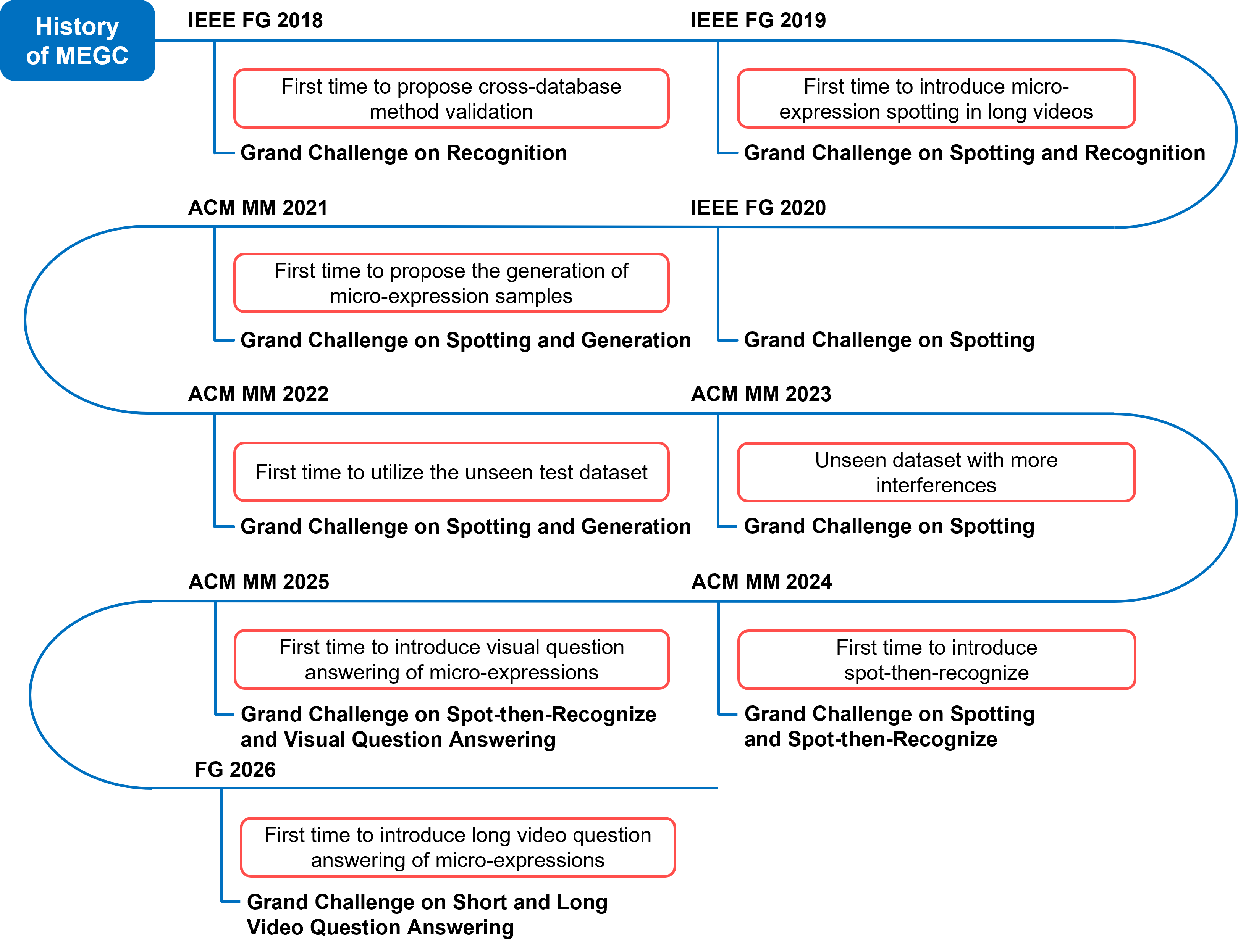}
    \caption{An overview of Micro-Expression Grand Challenges (MEGCs).}
    \label{fig:megc_history}
\end{figure*}
When a person attempts to suppress a facial expression, typically in a high-stakes scenario, there is a possibility of an involuntary movement occurring on the face, namely a facial micro-expression (ME)~\cite{ben2021video}. As such, the duration of an ME is very short, generally being no more than 500 milliseconds (ms), and is the telltale sign that distinguishes them from a normal facial expression~\cite{li2022cas}. Computational analysis and automation of tasks on MEs is an emerging area in multimedia research. However, only recently, the availability of a few spontaneously induced facial ME datasets has provided the impetus to advance further from the computational aspect.

Over the years, we have organised a series of ME grand challenges (MEGCs) to explore emerging directions and advance the state-of-the-art in ME research. To date, seven MEGCs have been held: FG’18~\cite{yap2018facial}, focusing on ME recognition; FG’19~\cite{see2019megc}, addressing both ME spotting and recognition; FG’20~\cite{li2020megc}, centered on ME spotting; MM’21~\cite{li2021fme}, introducing ME generation and continuing with spotting; MM’22~\cite{li2022megc2022}, continuing with spotting and generation; MM’23~\cite{davison2023megc2023}, with a focus on spotting; MM’24~\cite{see2024megc2024}, which introduced the spot-then-recognize paradigm and cross-cultural spotting; and MM’24~\cite{fan2025megc2025}, which introduced visual-question answering. An overview of all the MEGCs can be found in Fig.~\ref{fig:megc_history}.

This year marks the 9th MEGC, held in conjunction with Face and Gesture (FG) 2026. Two tasks are featured: (1) ME video question answering (ME-VQA), and (2) ME long-video question answering (ME-LVQA).

\textbf{Task 1: ME-VQA.} The rise of multimodal large language models (MLLMs) and large vision-language models (LVLMs) has opened new opportunities for ME analysis through their advanced multimodal reasoning capabilities. In this task, traditional ME annotations, such as emotion categories and action units, are reformulated into question-answer (QA) pairs. Given a video sequence of relatively short duration along with a natural language prompt, the model is expected to generate answers that describe the observed MEs, their attributes, and related information. This task introduces a novel multimodal direction for ME understanding, promoting interpretability, flexibility, and human-aligned interaction through natural language. ME-VQA was successfully introduced in MEGC 2025, and we continue with this task in MEGC 2026 with some modifications.

\textbf{Task 2: ME-LVQA.} ME-LVQA is a new task introduced in MEGC 2026. Extending from the ME-VQA task, ME-LVQA addresses the challenge of analyzing ME within long-duration video sequences in realistic, naturalistic settings. This task requires models to handle temporal reasoning over extended periods, spotting and understanding subtle MEs amidst natural facial movements and expressions. Participants must provide answers to questions about MEs in long videos with multiple expression events, making this a more challenging variant that tests the robustness and practical applicability of ME analysis methods in real-world scenarios. 

\section{Task 1: Video Question Answering}
\subsection{Task Formulation}
In the ME-VQA setting, the input to the model consists of (i) a sequence of video frames capturing subtle facial movements associated with MEs, and (ii) a corresponding natural language question related to the emotions or attributes of the sequence. The model is required to generate an answer in natural language, drawing on its understanding of both the visual cues and the question of MEs. These questions can cover a wide range of attributes, from binary classification such as ``Is the action unit lip corner depressor shown on the face?" to multiclass classification like ``What is the expression class?", and to more complex inquiries like ``Provide a detailed analysis of the micro-expression observed in this clip.".

\subsection{Dataset}
We do not have a restriction on the training set, while we recommend participants to use SAMM~\cite{davison2018samm}, CASME II~\cite{yan2014casme}, SMIC~\cite{li2013spontaneous}, CAS(ME)$^3$~\cite{li2022cas}, or 4DME~\cite{li20224DME} datasets for training.

To facilitate the task training, we curated a dataset from the SAMM, CASME II, and SMIC annotations to a ME-VQA (ME-VQA-v1) dataset\footnote{The ME-VQA version dataset is available at~\url{https://megc2025.github.io/challenge.html}}~\cite{fan2025megc2025}. In addition to this, in MEGC 2026, we refined some questions and answers and thus provide the ME-VQA-v2 dataset\footnote{The ME-VQA-v2 dataset is available at~\url{https://megc2026.github.io/challenge.html}} that is suitable to train an ME-VQA model.

The ME-VQA-v2 test set contains 24 ME clips, including 7 clips from the SAMM Challenge dataset \cite{yap20223d, davison2018samm} and 17 clips cropped from different videos in CAS(ME)$^3$ (unreleased before). The frame rate for SAMM is 200 fps, and the frame rate for CAS(ME)$^3$ is 30 fps. The participants should test on this unseen dataset.

\begin{table*}[!t]
    \footnotesize
    \centering
    \caption{Baseline results of the ME-VQA task with ME-VQA-v2 training and test sets. The test set results of the ME-VQA task using Qwen2.5VL-3B and Qwen3VL-4B (the baseline methods) trained on the curated ME-VQA dataset. The coarse emotion class refers to positive, negative, and surprise. The fine-grained emotion class refers to happiness, surprise, fear, disgust, anger, and sadness. ZS refers to zero-shot performance, while FT refers to fine-tuned performance.}
    \setlength{\tabcolsep}{3pt} 
    \begin{tabular}{@{}c@{\;}c@{\;}c@{\;}c@{\;}c@{\;}c@{\;}c@{\;}c@{\;}c@{\;}c@{\;}c@{\;}c@{\;}c@{\;}c@{\;}c@{\;}c@{\;}c@{\;}c@{\;}c@{\;}c@{}}
        \hline
        \multirow{4}{*}{Method} 
        & \multicolumn{6}{c}{SAMM} 
        & \multicolumn{6}{c}{CAS(ME)$^3$} 
        & \multicolumn{6}{c}{Overall} \\
        \cmidrule(lr){2-7} \cmidrule(lr){8-13} \cmidrule(lr){14-19} 
        & \multicolumn{2}{c}{Coarse} 
        & \multicolumn{2}{c}{Fine-Grained} 
        &  & 
        & \multicolumn{2}{c}{Coarse} 
        & \multicolumn{2}{c}{Fine-Grained} 
        &  & 
        & \multicolumn{2}{c}{Coarse} 
        & \multicolumn{2}{c}{Fine-Grained} 
        &  & 
        \\        
         & UF1 & UAR & UF1 & UAR & BLEU & ROUGE & UF1 & UAR & UF1 & UAR & BLEU & ROUGE & UF1 & UAR & UF1 & UAR & BLEU & ROUGE  \\
        \hline
        Qwen2.5VL-3B (ZS) & 0.242 & 0.333 & 0 & 0 & 0.090 & 0.295 & 0.262 & 0.333 & 0.057 & 0.048 & 0.089 & 0.315 & 0.256 & 0.333 & 0.057 & 0.048 & 0.065 & 0.312 \\
        Qwen2.5VL-3B (FT) & 0.206 & 0.25 & 0.057 & 0.1 & 0.234 & 0.331 & 0.376 & 0.515 & 0.044 & 0.111 & 0.109 & 0.375 & 0.328 & 0.422 & 0.059 & 0.117 & 0.122 & 0.362 \\
        \hline
        Qwen3VL-4B (ZS) & 0.242 & 0.333 & 0 & 0 & 0.068 & 0.229 & 0.262 & 0.382 & 0.114 & 0.222 & 0.065 & 0.382 & 0.256 & 0.333 & 0.079 & 0.200 & 0.066 & 0.339 \\
        Qwen3VL-4B (FT) & 0.242 & 0.333 & 0.178 & 0.250 & 0.128 & 0.416 & 0.262 & 0.333 & 0 & 0 & 0.236 & 0.413 & 0.256 & 0.333 & 0.052 & 0.064 & 0.204 & 0.415 \\
        \hline
    \end{tabular}
    \label{tab:vqa_results}
\end{table*}

\subsection{Evaluation Metrics}
To evaluate the performance of our ME-VQA baseline, we report metrics for both emotion classification and the overall language answer quality.

For both coarse- and fine-grained emotion classification, we use unweighted F1 Score (UF1) and Unweighted Average Recall (UAR) to ensure balanced evaluation across classes. The coarse emotion classes refer to positive, negative, and surprise. The fine-grained emotion classes refer to happiness, surprise, fear, disgust, anger, and sadness. The details of these metrics can be found in~\cite{see2019megc}.

For all VQA answers, we report bilingual evaluation understudy (BLEU)~\cite{papineni2002bleu} and recall-oriented understudy for gisting evaluation (ROUGE)-1~\cite{chin2004rouge} to assess the quality of generated text. BLEU evaluates n-gram precision between predicted and reference answers as:
\begin{equation}
    \mathrm{BLEU} = \exp \left( \min\left(1 - \frac{r}{c}, 0\right) + \sum_{n=1}^{N} w_n \log p_n \right),
\end{equation}
where $r$ is the reference length, $c$ is the candidate length, $p_n$ is the $n$-gram precision, and $w_n$ are the weights. 
The ROUGE-1 score is defined as the recall of unigram overlaps between the candidate answer $C$ and the reference answer $R$:
\begin{equation}
    \mathrm{ROUGE\text{-}1} = \frac{\sum_{w \in V} \min\big(\mathrm{N}_p(w),\ \mathrm{N}_r(w)\big)}{\sum_{w \in V} \mathrm{N}_r(w)},
\end{equation}
where $V$ is the set of unique words in the reference answer, and $\mathrm{N}_C(w)$ and $\mathrm{N}_R(w)$ are the occurrences of word $w$ in the candidate and reference answers, respectively.


\subsection{Base Method}
For the ME-VQA task, we employ the Qwen2.5VL-3B model~\cite{bai2025qwen25} as our baseline method. Qwen2.5VL-3B is a recent LVLM that is based on the Qwen 2.5 architecture and incorporates a vision encoder, a language model backbone, and a cross-modal fusion module. The model is pre-trained on large-scale image-text datasets, enabling it to perform a wide range of VQA tasks with strong zero-shot and few-shot generalisation capabilities.

To establish a general baseline, we consider zero-shot (ZS) and fine-tuning (FT) of the Qwen2.5VL-3B model. For fine-tuning, we applied QLoRA to the vision encoder, projection layers between vision and language, as well as the query and keys of the language model. All inputs to the models are videos.

\subsection{Baseline Results}
The results in Table~\ref{tab:vqa_results} indicate that both Qwen2.5VL-3B and Qwen3VL-4B exhibit limited performance on ME-VQA, particularly for fine-grained emotion recognition. In the zero-shot setting, both models achieve moderate coarse-level UF1/UAR (around 0.24–0.33), but fine-grained performance is extremely low, with UF1 close to zero in most cases. This suggests that while large VLMs possess some basic capability for coarse emotional discrimination, they struggle to capture subtle distinctions required for fine-grained ME classification.

Fine-tuning yields consistent but modest improvements, especially on CAS(ME)$^3$, where coarse-level UF1/UAR increases noticeably for Qwen2.5VL-3B. However, fine-grained recognition remains weak across datasets, indicating that subtle emotion differentiation remains a major challenge. In contrast, language quality metrics (BLEU and ROUGE) improve more clearly after fine-tuning, particularly for Qwen3VL-4B, suggesting that the models adapt more effectively in generating linguistically plausible answers than in achieving accurate ME classification. 


\section{Task 2: Long-Video Question Answering}

\subsection{Task Formulation}
ME-LVQA generalizes ME-VQA from short and pre-segmented clips to long video sequences that better reflect real-world facial behaviour analysis. Given a long video of spontaneous facial activity and a natural-language question, the goal of ME-LVQA is to generate an answer concerning the MEs present in the sequence, such as their occurrence, emotional categories, temporal characteristics, and associated action units. In contrast to ME-VQA settings, ME-LVQA formulation requires a model to unify temporal localisation and question answering within a single reasoning framework. The ME-LVQA task includes diverse question types designed to evaluate a comprehensive understanding of ME events in long videos. These questions cover event counting, such as the total number of expression, ME, or macro-expression (MaE) events, event categorisation, i.e., identifying whether a specific indexed event is ME or MaE, and attribute-level analysis, i.e., listing distinct action units that occur throughout the video. Such questions require models to reason over the entire temporal sequence, distinguish between different expression types, aggregate information across multiple events, and provide structured yet natural-language answers grounded in long-range video evidence. Participants must use MLLM or VLMs to tackle this task.

\begin{table*}[t]
    \centering
    \caption{Baseline results of the ME-LVQA task. The test set results of the ME-LVQA task using Qwen2.5VL-3B and Qwen3VL-4B (the baseline methods) trained on the curated ME-LVQA dataset. The fine-tuning used 5 subjects from SAMM-LV and 5 subjects from CAS(ME)$^3$. ZS refers to zero-shot performance, while FT refers to fine-tuned performance.}
    \label{tab:results}
    \textbf{(a) SAMM}\\
    \vspace{2pt}  
    \begin{tabular}{l cc cc cc cc cc}
    \toprule
    & \multicolumn{2}{c}{\textbf{Expression}} & \multicolumn{2}{c}{\textbf{ME}} & \multicolumn{2}{c}{\textbf{MaE}} & \multicolumn{2}{c}{\textbf{AU}} & \multicolumn{2}{c}{\textbf{Expression Type}} \\
    \cmidrule(lr){2-3} \cmidrule(lr){4-5} \cmidrule(lr){6-7} \cmidrule(lr){8-9} \cmidrule(lr){10-11}
    \textbf{Method} & MAE & RMSE & MAE & RMSE & MAE & RMSE & F1 & Jaccard & UF1 & UAR \\
    \midrule
    Qwen2.5VL-3B (ZS) & 5.80 & 6.48 & 3.10 & 3.89 & 2.10 & 2.70 & 0.161 & 0.098 & 0.373 & 0.490 \\
    Qwen2.5VL-3B (FT) & 5.20 & 6.05 & 3.20 & 4.02 & 2.10 & 2.70 & 0.169 & 0.110 & 0.560 & 0.636 \\
    Qwen3VL-4B (ZS)   & 6.20 & 6.84 & 3.60 & 4.27 & 3.10 & 3.54 & 0.075 & 0.048 & 0.479 & 0.525 \\
    Qwen3VL-4B (FT)   & 5.50 & 6.35 & 3.30 & 4.11 & 3.00 & 3.44 & 0.167 & 0.103 & 0.487 & 0.591 \\
    \bottomrule
    \end{tabular}
    
    \vspace{2pt}
    \textbf{(b) CAS(ME)$^3$}\\
    \vspace{2pt}
    \begin{tabular}{l cc cc cc cc cc}
    \toprule
    & \multicolumn{2}{c}{\textbf{Expression}} & \multicolumn{2}{c}{\textbf{ME}} & \multicolumn{2}{c}{\textbf{MaE}} & \multicolumn{2}{c}{\textbf{AU}} & \multicolumn{2}{c}{\textbf{Expression Type}} \\
    \cmidrule(lr){2-3} \cmidrule(lr){4-5} \cmidrule(lr){6-7} \cmidrule(lr){8-9} \cmidrule(lr){10-11}
    \textbf{Method} & MAE & RMSE & MAE & RMSE & MAE & RMSE & F1 & Jaccard & UF1 & UAR \\
    \midrule
    Qwen2.5VL-3B (ZS) & 16.35 & 19.67 & 8.35 & 11.05 & 5.95 & 7.10 & 0.189 & 0.110 & 0.355 & 0.500 \\
    Qwen2.5VL-3B (FT) & 12.15 & 15.58 & 9.35 & 12.56 & 5.95 & 7.10 & 0.152 & 0.099 & 0.400 & 0.505 \\
    Qwen3VL-4B (ZS)   & 16.65 & 19.82 & 10.20 & 13.17 & 6.80 & 7.84 & 0.205 & 0.132 & 0.481 & 0.538 \\
    Qwen3VL-4B (FT)   & 15.85 & 19.25 & 9.80 & 12.73 & 6.80 & 7.80 & 0.246 & 0.154 & 0.355 & 0.500 \\
    \bottomrule
    \end{tabular} 
    
    \vspace{2pt}
    \textbf{(c) Overall (CAS(ME)$^3$ + SAMM)}\\
    \vspace{2pt}
    \begin{tabular}{l cc cc cc cc cc}
    \toprule
    & \multicolumn{2}{c}{\textbf{Expression}} & \multicolumn{2}{c}{\textbf{ME}} & \multicolumn{2}{c}{\textbf{MaE}} & \multicolumn{2}{c}{\textbf{AU}} & \multicolumn{2}{c}{\textbf{Expression Type}} \\
    \cmidrule(lr){2-3} \cmidrule(lr){4-5} \cmidrule(lr){6-7} \cmidrule(lr){8-9} \cmidrule(lr){10-11}
    \textbf{Method} & MAE & RMSE & MAE & RMSE & MAE & RMSE & F1 & Jaccard & UF1 & UAR \\
    \midrule
    Qwen2.5VL-3B (ZS) & 12.83 & 16.49 & 6.60 & 9.30 & 4.67 & 6.00 & 0.180 & 0.106 & 0.369 & 0.501 \\
    Qwen2.5VL-3B (FT) & 9.83 & 13.19 & 7.30 & 10.51 & 4.67 & 6.00 & 0.158 & 0.102 & 0.467 & 0.553 \\
    Qwen3VL-4B (ZS)   & 13.17 & 16.66 & 8.00 & 11.03 & 5.57 & 6.72 & 0.161 & 0.104 & 0.487 & 0.539 \\
    Qwen3VL-4B (FT)   & 12.40 & 16.14 & 7.63 & 10.66 & 5.53 & 6.67 & 0.219 & 0.137 & 0.413 & 0.535 \\
    \bottomrule
    \end{tabular}
    \label{tab:lvqa_results}
\end{table*}

\subsection{Dataset}
We do not have a restriction on the training set. For the ME-LVQA task, participants may use any of the standard ME long-video datasets, such as SAMM-LV~\cite{yap2020samm}, CAS(ME)$^3$~\cite{li2022cas}, CAS(ME)$^2$, 4DME~\cite{li20224DME}, or other long-duration naturalistic video datasets. 

To facilitate the task training, we curated a ME-LVQA training set\footnote{The ME-LVQA dataset is available at~\url{https://megc2026.github.io/challenge.html}} that contains SAMM-LV and CAS(ME)$^3$, and added the relevant QA pairs. 

The test set of ME-LVQA, it contains 30 long videos, including 10 long videos from the SAMM Challenge dataset \cite{yap20223d, davison2018samm} and 20 clips cropped from different videos in CAS(ME)$^3$ (unreleased before). The frame rate for SAMM Challenge dataset is 200fps, and the frame rate for CAS(ME)$^3$ is 30 fps. The participants should test on this unseen dataset.

\subsection{Evaluation Metrics}
To evaluate the performance of the ME-LVQA task, we adopt task-specific metrics according to the nature of each question type. 

For questions that require predicting the number of expression events, including the total number of expression events as well as the numbers of MEs and MaEs, we formulate the problem as a regression task. We report mean absolute error (MAE) and root mean squared error (RMSE).  MAE measures the average magnitude of the prediction error, while RMSE penalizes larger deviations more strongly. They are defined as:
\begin{equation}
\mathrm{MAE} = \frac{1}{N}\sum_{i=1}^{N} |\hat{y}_i - y_i|,
\end{equation}
\begin{equation}
\mathrm{RMSE} = \sqrt{\frac{1}{N}\sum_{i=1}^{N} (\hat{y}_i - y_i)^2},
\end{equation}
where $y_i$ and $\hat{y}_i$ denote the ground-truth and predicted number of events for the $i$-th sample, respectively, and $N$ is the number of evaluated samples.

For questions asking about the distinct AU appearing in the video, the output is a set of AU descriptions. Since the order of AUs in the answer does not affect correctness, we treat this as a set prediction problem and evaluate it using set-based F1 score and the Jaccard index. These metrics measure the overlap between the predicted and reference AU sets and are insensitive to the ordering of AU descriptions.
Given the predicted AU set $\hat{A}$ and the ground-truth AU set $A$, the set-based F1 score is defined as:
\begin{equation}
F1_{\text{AU}} = \frac{2|\hat{A} \cap A|}{|\hat{A}| + |A|},
\end{equation}
and the Jaccard index is defined as:
\begin{equation}
\mathrm{Jaccard}_\text{AU} = \frac{|\hat{A} \cap A|}{|\hat{A} \cup A|}.
\end{equation}

For questions asking about the type of a specific expression event (ME or MaE), the problem is treated as a binary classification task. Following the evaluation protocol commonly used in ME analysis, we report unweighted F1 score (UF1) and unweighted average recall (UAR)~\cite{see2019megc}. 

\subsection{Baseline Method}
For the ME-LVQA task, we employ Qwen2.5VL-3B~\cite{bai2025qwen25} and Qwen3VL-4B~\cite{bai2025qwen3} as baseline models. All the inputs to the models are videos. Qwen2.5VL-3B is a compact VLM model that incorporates a vision encoder, language model backbone, and cross-modal fusion module, while Qwen3VL-4B represents an enhanced variant with improved multimodal reasoning capabilities. 

Both models are evaluated in zero-shot (ZS) and fine-tuned (FT) settings to assess their effectiveness on long video ME reasoning. For the fine-tuning experiments, we adopt quantized low-rank adaptation (QLoRA) to efficiently update model parameters. Specifically, we apply LoRA adapters to the vision encoder and the language model components. The fine-tuning is conducted for 5 epochs using a reduced and curated subset from ME-LVQA training data, comprising randomly selected 5 subjects from SAMM and 5 subjects from CASME3. The selection is due to limited computing resources. 


\subsection{Baseline Results}
Table~\ref{tab:lvqa_results} presents the baseline evaluation results for ME-LVQA across the SAMM and CAS(ME)$^3$ challenge test set, and their combination. The results reveal significant performance degradation compared to the short-clip ME-VQA task, highlighting the substantial challenges introduced by long-duration video understanding.

The results confirm that long-video ME understanding is substantially more challenging than short-clip analysis. Although fine-tuning improves performance, it does not close the gap in ME counting and AU recognition, where errors remain high. Temporal localisation and fine-grained facial action modelling, therefore, remain major bottlenecks. 

These findings should be interpreted in light of the limited fine-tuning setup, which used only 10 subjects (5 from SAMM and 5 from CAS(ME)$^3$). Such restricted subject diversity likely limits identity-invariant learning and generalisation to unseen individuals. Given the person-specific and subtle nature of MEs, the modest gains suggest that models may be learning partial subject-dependent patterns rather than robust temporal dynamics, underscoring the need for larger-scale and more diverse training data for the participants to explore.

{\small
\bibliographystyle{ieee}
\bibliography{egbib}
}

\end{document}